\documentclass[11pt]{article}
\usepackage{fa2025}
\usepackage{amsmath}
\usepackage{cite}
\usepackage{url}
\usepackage{graphicx}
\usepackage{color}
\usepackage{siunitx}
\usepackage[utf8]{inputenc}
\usepackage[colorinlistoftodos]{todonotes}

\title{Clustering and novel class recognition: evaluating bioacoustic deep learning feature extractors}

\multauthor{
  Vincent S. Kather$^{1, 2*}$ 
  \hspace{1cm} 
  Burooj Ghani$^2$ 
  \hspace{1cm} 
  Dan Stowell$^{1, 2}$
  } { 
  $^1$ Department of Cognitive Science and Artificial Intelligence, Tilburg University, Netherlands\\
  $^2$ Naturalis Biodiversity Center, Leiden, Netherlands\\
  \correspondingauthor{vincent.kather@naturalis.nl}{Vincent S. Kather et al.}
}

\begin{document}

\maketitle
\begin{abstract}
% In computational bioacoustics, deep learning models are composed of feature extractors and classifiers. 
% The feature extractors generate vector representations of the input sound segments, called embeddings. 
% The classifiers recognize a fixed number of classes most commonly representing different animal species. 
% Various benchmarks have been published to evaluate the classification performance. 
% While such benchmarking provides insight into specific performance statistics, it is limited to species that are included in the models training data. Furthermore, it makes it impossible to compare models trained on very different taxonomic groups. 
% This paper aims to address this gap by analyzing the generated embeddings of more than 15 bioacoustic models spanning a wide range of setups (model architectures, training data, training paradigms). 
% We investigate and evaluate different ways to quantify how models structure embedding spaces, which allows us to focus our comparison on feature extractors independent of classifiers. 
% We believe that this approach lets us evaluate the adaptability and generalization potential of models going beyond the classes they were trained on.
In computational bioacoustics, deep learning models are composed of feature extractors and classifiers. 
The feature extractors generate vector representations of the input sound segments, called embeddings, which can be input to a classifier. 
% The classifiers recognize a fixed number of classes most commonly representing different animal species. 
% Various benchmarks have been published to evaluate the classification performance. 
While benchmarking of classification scores provides insights into specific performance statistics, it is limited to species that are included in the models' training data. Furthermore, it makes it impossible to compare models trained on very different taxonomic groups. 
This paper aims to address this gap by analyzing the embeddings generated by the feature extractors of 15 bioacoustic models spanning a wide range of setups (model architectures, training data, training paradigms). 
We evaluated and compared different ways in which models structure embedding spaces through clustering and kNN classification, which allows us to focus our comparison on feature extractors independent of their classifiers. 
We believe that this approach lets us evaluate the adaptability and generalization potential of models going beyond the classes they were trained on.
\end{abstract}


\section{Introduction}
\label{sec:introcution}

Human-driven climate change and deforestation have caused a rapid decline in global biodiversity \cite{gosselin_relationships_2018}.
Using sensor arrays for biodiversity monitoring, ecologists can gather information on environments and investigate how human pressures affect biodiversity and how we can halt its decline \cite{schmeller_building_2017}.
Passive acoustic monitoring (PAM) is one method that provides a low-cost and non-invasive way to monitor biodiversity \cite{sugai_terrestrial_2019}.
The vast amount of data generated by PAM sensors has led to the rapid development of bioacoustic deep learning models to help researchers reduce the annotation effort \cite{stowell_computational_2022}.
The use of these state-of-the-art models has proven valuable in ecological studies, for example, in general species assessments \cite{cowans_improving_2024} or detection of endangered species \cite{allen-ankins_use_2025}.

While bioacoustic deep learning models are a useful asset in ecology, it is crucial to understand the models' limitations based on their training setup.
% little is known about how their training setup correlates to their performance.
The two main training strategies to develop bioacoustic deep learning models are supervised learning and self-supervised learning.
% In supervised learning large acoustic datasets are collected, annotated and used to train \cite{kahl_birdnet_2021}. 
Supervised learning models require large amounts of annotated data to be trained \cite{hagiwara_aves_2022}.
The models classify sounds based on a fixed number of predefined classes representing annotated sounds in the dataset.
While supervised learning can have the benefit of instructing models to differentiate between species vocalizations, it requires annotated datasets.
This limits supervised learning models to known and annotated classes and makes them sensitive to class imbalance and label quality.
% In self-supervised learning classes are defined by the model during training time \cite{baevski_efficient_2023,huang_masked_2022}.
% Self-supervised learning can be executed in different ways.
Alternatively, self-supervised learning can be trained without annotated data, enabling models to learn meaningful representations directly from the data.
For example, in a paradigm called masked prediction, the model is trained to predict a masked portion of the audio, thus modeling bioacoustic characteristics without supervision \cite{huang_masked_2022}.
With growing annotated databases, supervised learning models like BirdNET \cite{kahl_birdnet_2021} improve in performance, yet recent developments in self-supervised learning on models like Animal2vec \cite{schafer-zimmermann_animal2vec_2024} indicate a promising new direction for the field requiring less manual annotation.
Not requiring annotations, self-supervised learning models can be trained on far larger datasets, however there is no control of what is being learned.
The model might, for instance, learn to differentiate between sounds based on characteristics very different from the species that produces them.

% The datasets that are created in this manner tend to feature many annotations for few classes and few annotations for many classes (long-tail distribution/class imbalance) \cite{arnaud_improving_2023}. 
% Models with a heavily imbalanced training dataset have shown to yield performance imbalances \cite{hamer_birb_2023}. 
% This implies that the models results are only reproducible when used to classify the well represented majority classes. 
% Furthermore, models trained in a supervised learning pipeline are restricted to only recognize a fixed set of classes. 
% Class imbalance and closed set recognition are two examples of phenomena, that show while supervised learning is the convention, it has shortcomings that restrict the usability of models trained in this way.

To evaluate bioacoustic deep learning models, a comparison of just the classifier performance obscures the fine-grained differences between models and the way they analyze input sounds.
Bioacoustic deep learning models consist of artificial neural networks that can be subdivided into feature extractors and classifiers. 
The feature extractor creates an embedding (vector representation) of an input sound and the classifier (which corresponds to the final dense layer of the model) maps the embedding onto classes.
Commonly, in bioacoustics, a suite of established benchmarks is used to compare the classifier performance of state-of-the-art models \cite{hamer_birb_2023}.
This requires the models to have been trained on the classes present in the benchmarking datasets.
However, there is an alternative: using the embeddings created by the feature extractors, the embedding spaces can be analyzed with regard to their structural characteristics, irrespective of what the classifiers were trained on.

%Output dimensions of feature extractors vary greatly, but it is uncertain if their dimensionality correlates with the downstream classifier performance. 
%Dimensionality reduction algorithms are useful to standardize the dimensionality of different feature extractors, as well as help visualize the high dimensional embedding spaces.
% Due to the lack of separable clusters in ecoacoustics and soundscape analyzes, evaluation of embedding spaces (reduced to two dimensions) has become common in recent years \cite{sethi_characterizing_2020,calonge_revised_2024,parcerisas_categorizing_2023}. 
%However, it is not established if reducing the dimentionality of the embedding space has an effect on the performance of downstrem tasks.
%We therefore compare performance in both original and reduced embedding spaces.

Due to their size, datasets that are used to train bioacoustic models for bird detection are often based on citizen science data (e.g., xeno-canto \cite{xeno-canto_xeno-canto_2025}).
The majority of these recordings are focal recordings of individuals which are weakly labeled with little polyphony.
When models are applied to large PAM datasets, especially outside of North America (where the majority of the training data originates), the difference in recording conditions causes performance drops \cite{perez-granados_birdnet_2023}.
To accurately evaluate the models in this study, we use a PAM bird vocalization dataset from Colombia and Costa Rica \cite{vega-hidalgo_collection_2023}, as well as a dataset of frog vocalizations from Brazil \cite{canas_dataset_2023}.
Both datasets are comprised of PAM recordings in noisy environments and can therefore be considered as challenging datasets.
With this, we hope to on the one hand emphasize the performance differences between the models and on the other hand produce results that will reflect in real world applications.

This study aims to showcase the potential for evaluating and especially comparing bioacoustic deep learning models with regard to their training paradigm and training data.
The evaluation focuses on the structuring capabilities of each feature extractor, as measured by clustering and classification performance.
Classification in this case refers to recognition of novel classes, as none of the models have been pretrained on the classes selected here.
This way, classification heads are attached to each feature extractor, all of which are trained on the same evaluation sets (bird and frog vocalizations), i.e. same classes and same data, allowing us to compare their performance.
Classification is done using a k-nearest neighbor (kNN) approach.
% We perform our analysis in both the original embedding space and a reduced dimensional embedding space. 
% That way we ensure the dimension is standardized for the second evaluation, whilst we can investigate performance differences between the original embedding space and the reduced space.
% This study will therefore examine how different dimensionality reduction strategies impact performance. 
% While the possibilities of analysis of these high dimensional embedding spaces are vast, we limit ourselves to 
% Performance will be evaluated through an analysis of clustering based on
% \begin{enumerate}
%     \item Silhouette Score \cite{rousseeuw_silhouettes_1987}, 
%     \item Adjusted Rand Index \cite{steinley_variance_2016} based on KMeans clustering and
%     \item Adjusted Mutual Information \cite{romano_standardized_2014} based on KMeans clustering.
% \end{enumerate}
% To do so, we analyse clusterings of deep learning feature extractors and use them to compare different training setups.
This method of analysis opens up the possibilities for a fair comparison of deep learning feature extractors guiding the field to a better understanding of how training configurations affect downstream performance.
% Summary of what we are contributing.

\section{Methods}
\label{sec:methods}

\begin{table*}[t]
  
  \caption{List of feature extractors compared in this study. Columns "abbrev." shows the an abbreviated name used in Fig. \ref{fig:orig_vs_ump}. "training" shows the training setup chosen during training, i.e. ssl for self-supervised learning, supl for supervised learning and ft for fine-tuning. The "architecture" column more specifically describes the model architecture used. "dimension" shows the output dimension of the feature extractor. "trained on" summarizes the training data of the model. "ref." provides the respective publication.}
  \label{tab:bacpipe_models}

  \centering
  \begin{tabular}{l|c|c|c|c|l|c}
     \hline
    name& abbrev. & 
    training & architecture &
    dimension & trained on & ref.\\
      %  ref paper&
      %  ref code&
      %  sampling rate&
    \hline
    Animal2vec\_XC      & a2v\_xc   & ssl & d2v2.0 & 768 & birds & \cite{schafer-zimmermann_animal2vec_2024}\\
    Animal2vec\_MK      & a2v\_mk  & ssl + ft & d2v2.0 & 1024& meerkats & \cite{schafer-zimmermann_animal2vec_2024}\\
    AudioMAE            & aud\_mae  & ssl + ft & ViT 	 & 768 & general & \cite{huang_masked_2022}\\
    AVES\_ESpecies      & aves   & ssl + ft & HuBERT 	 & 768 & general + animals & \cite{hagiwara_aves_2022}\\
    AvesEcho\_PaSST     & aecho   & supl & PaSST 	 & 768 & birds & \cite{ghani_generalization_2024}\\
    BioLingual          & bioling  & supl & CLAP 	 & 512 & animals + birds & \cite{robinson_transferable_2023}\\
    BirdAVES\_ESpecies  & birdaves   & ssl + ft & HuBERT 	 & 1024& general + birds & \cite{hagiwara_aves_2022}\\
    BirdNET             & brdnet   & supl & EffNetB0 	 & 1024& birds & \cite{kahl_birdnet_2021}\\
    Google\_Whale       &  g\_whale  & supl & EffNetB0 	 & 1280& whales & - \\
    Insect459NET        & i459 & supl & EffNetv2s 	 & 1280& insects & - \\
    Insect66NET         & i66 & supl & EffNetv2s 	 & 1280& insects & - \\
    NonBioAVES\_ESpecies& nonbioaves    & ssl + ft & HuBERT 	 & 1024& general + non-bio & \cite{hagiwara_aves_2022}\\
    Perch\_Bird         & perch    & supl & EffNetB0 	 & 1280& birds & - \\
    ProtoCLR            & p\_clr   & supl 	 & CvT-13 & 384 & birds & \cite{moummad_domain-invariant_2024}\\
    SurfPerch           &  s\_perch  & supl & EffNetB0 	 & 1280& coral reefs + birds & \cite{williams_leveraging_2024}\\
    % \hline
  \end{tabular}
\end{table*}

To incorporate a variety of training setups, covering popular models as well as models targeted to various species groups, we compare a total of 15 pretrained bioacoustic feature extractors (Table \ref{tab:bacpipe_models}).
This includes both self-supervised and supervised learning feature extractors. 
Furthermore, large variations in input length, embedding dimension and training data provide a diverse landscape of feature extractors, allowing us to analyze performance of differently structured embedding spaces.

\subsection{Dataset}
\label{ssub:dataset}

The evaluation datasets that were used for this study are a bird vocalization dataset (bird dataset) recorded in coffee farms in Colombia and Costa Rica \cite{vega-hidalgo_collection_2023} and a frog vocalization dataset (frog dataset) recorded in Brazil \cite{canas_dataset_2023}.
The recordings in these datasets feature challenging soundscape recordings with overlapping vocalizers and noisy environments.
Both datasets have been reduced from their original size to only include sound events corresponding to classes with more than 150 annotations.
Due to the high amount of polyphony, the frog dataset has furthermore been reduced to only contain sound events with non-overlapping annotations, i.e. turning it into a single-label dataset.
It is worth mentioning, that by excluding overlapping annotations, only polyphony of frog species is removed, the complex background of overlapping sounds created by insects and birds in the recordings remain.

For the bird dataset this results in 11 classes, while for the frog dataset 18 classes are included in the final dataset.
We intentionally selected soundscape recordings with gradual changes of background noise and overlapping species vocalizations to amplify the differences between the feature extractors' capabilities to structure the data.

\subsection{Data pipeline}
\label{ssub:data_pipe}

For each of the feature extractors, the respective model code base was cloned, and the model was stripped of its classifier.
For both animal2vec feature extractors, outputs from the attention heads and input lengths are averaged, resulting in one embedding per input segment (as is the case with all other feature extractors).
Data are imported from the sound files, resampled to the model-specific sample rate, and padded to fit the model-specific input length.
All the necessary code to reproduce the computations can be found in the repository \textbf{bacpipe}\footnote{\url{github.com/bioacoustic-ai/bacpipe}} (\textbf{b}io\textbf{a}coustic \textbf{c}ollection \textbf{pipe}line).

\subsection{Methods for evaluating embedding spaces}
\label{ssub:eval_dim_reduc}

Embedding spaces are evaluated using clustering and classification.
Our primary focus is the comparison of the two paradigms: supervised learning and self-supervised learning.
Furthermore, we explore how the data chosen for training affects the clustering capabilities of different feature extractors.

The clustering is computed using K-Means with the same number of clusters as classes in the ground truth. 
Clustering performance is evaluated using Adjusted Mutual Information (AMI) \cite{romano_standardized_2014} to compare the K-Means clustering with the ground truth.
% Adjusted Rand Index is not included in this study, as it focuses on how well data points are grouped in a clustering, whereas we are primarily interested how well the KMeans clustering agrees with the ground truth labels.
% Non-linear dimensionality reduction algorithms like UMAP create a learned transformation based on a graph structure of the data and therefore disregards distances between data points \cite{mcinnes_umap_2020}.
% Silhouette Score is also not included in this comparison as the challenging dataset yielded very low performance and variance, making a meaningful comparison impossible.
% AMI and ARI require a clustering to be computed, to then quantify the agreement between that clustering and the ground truth and are therefore applicable to non-linear dimensionality reductions.
% AMI measures how well clusters share information while ARI quantifies how well points are grouped.

Performance is also evaluated by training a simple kNN classifier on each of the embedding spaces with a nearest neighbor parameter of 15.
% The linear classifier is trained on the 11 and 18 classes for the bird and frog datasets respectively for 10 epochs with a batch size of 64 and a learning rate of 0.001. 
% For the kNN classifier a nearest neighbor parameter of 15 is chosen.
The classifier is trained on the 11 and 18 classes for the bird and frog datasets respectively.
Data are split into train, validation and test set in the ratio 0.65:0.15:0.2.
Performance is evaluated using a balanced macro accuracy score \cite{brodersen_balanced_2010} to handle the imbalance in class sizes.

Evaluations are computed in both the original embedding spaces and an embedding space reduced to 300 dimensions.
This way the embedding dimension is standardized and performance can be compared while controlling for this factor.
% It also allows us to compare the performance of each model in their high dimensional original embedding space, as well as in a reduced dimension.
% For the linear classifier, to preserve relative distances between data points, Principal Component Analysis (PCA), a linear dimensionality reduction is selected.
Uniform manifold approximation projection (UMAP) \cite{mcinnes_umap_2020} is selected for the dimensionality reduction to 300 dimensions.
UMAP is also used to visualize the embeddings in two dimensions in Fig. \ref{fig:embeds}.

\section{Results}
\label{sec:results}

\begin{table}[t]
  
  \caption{Classification (macro accuracy) and clustering (AMI) performance of the original embeddings, and UMAP embeddings reduced to 300 dimensions. Values show the mean over all ssl, supl, bird and non-bird feature extractors, corresponding to the categories and colors in Fig.~\ref{fig:orig_vs_ump}. Best values among all categories and dimensionalities are bold.}
  \label{tab:results}
  \centering
  \begin{tabular}{l|cc|cc}
    \hline
    \textbf{bird data}& \multicolumn{2}{c|}{classification} & \multicolumn{2}{c}{clustering}\\
    % \hline
    category &
    original &
    UMAP &
    original &
    UMAP \\
    \hline
    supl & 0.695 & {0.711} & 0.418 & {0.476}\\
    ssl & 0.617 & {0.619} & 0.256 & {0.405}\\
    bird & 0.712 & {\textbf{0.723}} & 0.426 & {\textbf{0.479}}\\
    non-bird & {0.658} & 0.618 & 0.271 & {0.413}\\
    % \hline
    % \hline
    \textbf{frog data}& &&& \\
    % \multicolumn{2}{c|}{classification} & \multicolumn{2}{c}{clustering}\\
    % \hline
    category &
    original &
    UMAP &
    original &
    UMAP \\
    \hline
    supl & {0.668} & 0.655 & 0.488 & {0.546}\\
    ssl & {0.682} & 0.679 & 0.414 & {0.508}\\
    bird & {\textbf{0.705}} & 0.698 & 0.5 & {\textbf{0.558}}\\
    non-bird & 0.638 & 0.627 & 0.414 & {0.5}\\
    % \hline

    % \hline
  \end{tabular}
\end{table}

Table \ref{tab:results} shows the averaged performances over all feature extractors, grouped into the different categories: supervised vs. self-supervised learning, and trained on bird data vs. non-bird data.
Firstly, we want to point out the performance changes following dimensionality reduction using UMAP.
For evaluation with bird and frog datasets, UMAP embeddings significantly improve clustering results for each category.
For the classification performance, values are very similar between original and UMAP reduced embeddings, however, when applied to the bird data, UMAP yields performance increases for most categories, whereas for the frog data, the original embeddings yield better results.

\begin{figure}[ht]
    \centerline{{
    \includegraphics[width=8.1cm]{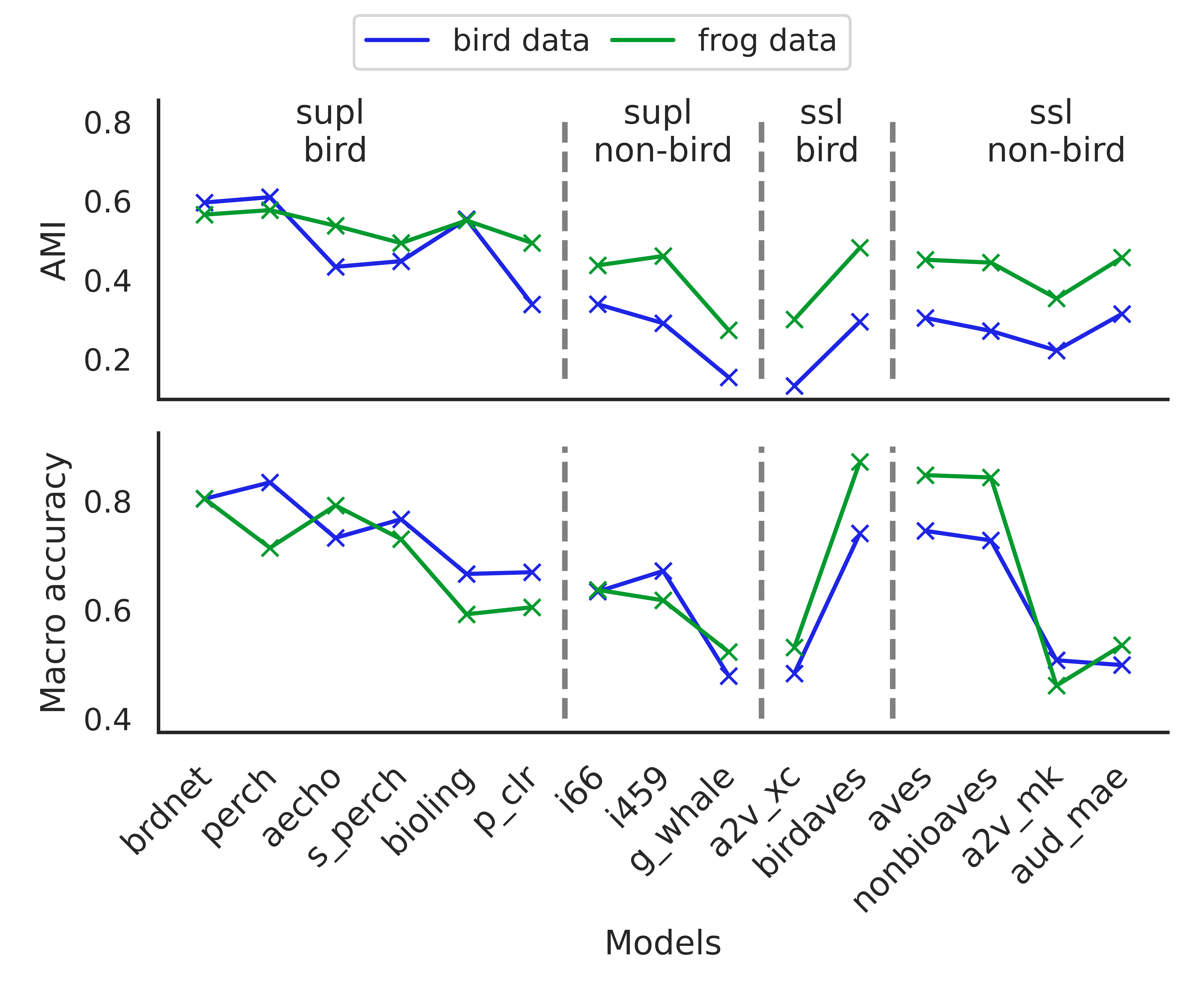}}}
    \caption{Comparison of feature extractor performance by learning paradigm, training data and application data. 
    The top plot shows clustering results of AMI.
    The bottom plot shows macro accuracy results of kNN classification. 
    Colors correspond to performance of the feature extractors when applied to bird data (blue) and frog data (green).
    The x-axis shows abbreviated model names, corresponding to "abbrev." column in Tab.~\ref{tab:bacpipe_models}. 
    Models are grouped into categories along the x-axis by training paradigm (supervised learning (supl) and self-supervised learning (ssl)) and training data (bird data and non-bird data, see Table~\ref{tab:bacpipe_models}).}
    \label{fig:orig_vs_ump}
\end{figure}

Supervised feature extractors outperform self-supervised ones by a large margin for classification and clustering on the bird dataset.
For the frog dataset, due to the improved performance of the AVES models (see Fig. \ref{fig:orig_vs_ump}), self-supervision outperforms supervised learning measured by kNN classification.
When comparing the values between all categories, the bird-trained feature extractors outperform all other categories for both bird and frog evaluation sets.

% Embeddings spaces of are visualized using UMAP in are generated from the input data using all feature extractors and then the dimensionality reduction algorithm UMAP is used to visualize the data in two dimensions.
% The results are shown in Fig. \ref{fig:embeds}.

% To highlight performance changes once the feature extractors are applied to a dataset different from their training domain, Fig. \ref{fig:orig_vs_ump} shows the performance on the bird dataset (blue) and the frog dataset (green).
Figure \ref{fig:orig_vs_ump} shows the averaged performances for each of the feature extractors.
First, we will focus on the feature extractors that are applied to the bird dataset, shown in blue.
When focusing on the clustering (top), the six best performing feature extractors are trained using supervised learning, while all self-supervised learning feature extractors reach clustering performances under 0.31.
Performance measured by kNN classification is more mixed, however again supervised learning feature extractors reach the three highest values.
Furthermore, Animal2vec\_XC, the only self-supervised learning feature extractor that was not fine-tuned, performs poorly for both clustering and kNN classification.
Google\_Whale represents the only supervised learning feature extractor performing very poorly by clustering and kNN classification.

Comparing by training data, feature extractors trained on only or including bird datasets (supl bird and ssl bird) outperform the other feature extractors by kNN classification and even more so by clustering, with the exception of Animal2vec\_XC.
Aside from ProtoCLR the supervised learning models that are also trained on birds (supl bird) vastly outperform all other models by clustering and with the exception of the AVES feature extractors also by kNN classification.
Perch and BirdNET lead both in clustering and kNN classification by a large margin.
Biolingual, which was trained on large bird databases using a multi-modal approach performs well by clustering, but comparatively poorly by kNN classification.

In green we can see the performance when the feature extractors are applied to the frog data.
Among supervised learning feature extractors, changes in both clustering and kNN classification performance are varied.
Especially the three best performing models by clustering for the bird data, Perch, BirdNET and Biolingual, all decrease in performance when applied to the frog dataset.
The non-bird trained supervised learning feature extractors (supl non-bird) all improve their clustering performance by a large margin.
All self-supervised feature extractors vastly improve in clustering performance and aside from Animal2vec\_MK also in kNN classification.
The self-supervised AVES feature extractors (BirdAVES, AVES and NonBioAVES) drastically improve in both clustering and kNN classification on the frog data.
So much so, that all three outperform all other feature extractors by kNN classification.
% Aside from all feature extractors increase in clustering performance.

All non-bird trained feature extractors improve in clustering when applied to frog data.
Again, when applied to the frog dataset, all bird trained feature extractors except for Animal2vec\_XC outperform the rest by clustering and the top six models are the supervised learning bird trained feature extractors.
Among them, only AvesEcho improves in both clustering and kNN classification.
However, by kNN classification, results are more mixed by both training paradigm and training domain.

% When referring back to Table \ref{tab:bacpipe_models} embedding dimension does not correlate with clustering or kNN classification performance.
% Furthermore, the only two feature extractors trained on marine sounds, Google\_Whale and SurfPerch (trained on birds and marine sounds) reach very different performances.

\begin{figure*}[ht]
    \centerline{{
    \includegraphics[width=16.3cm]{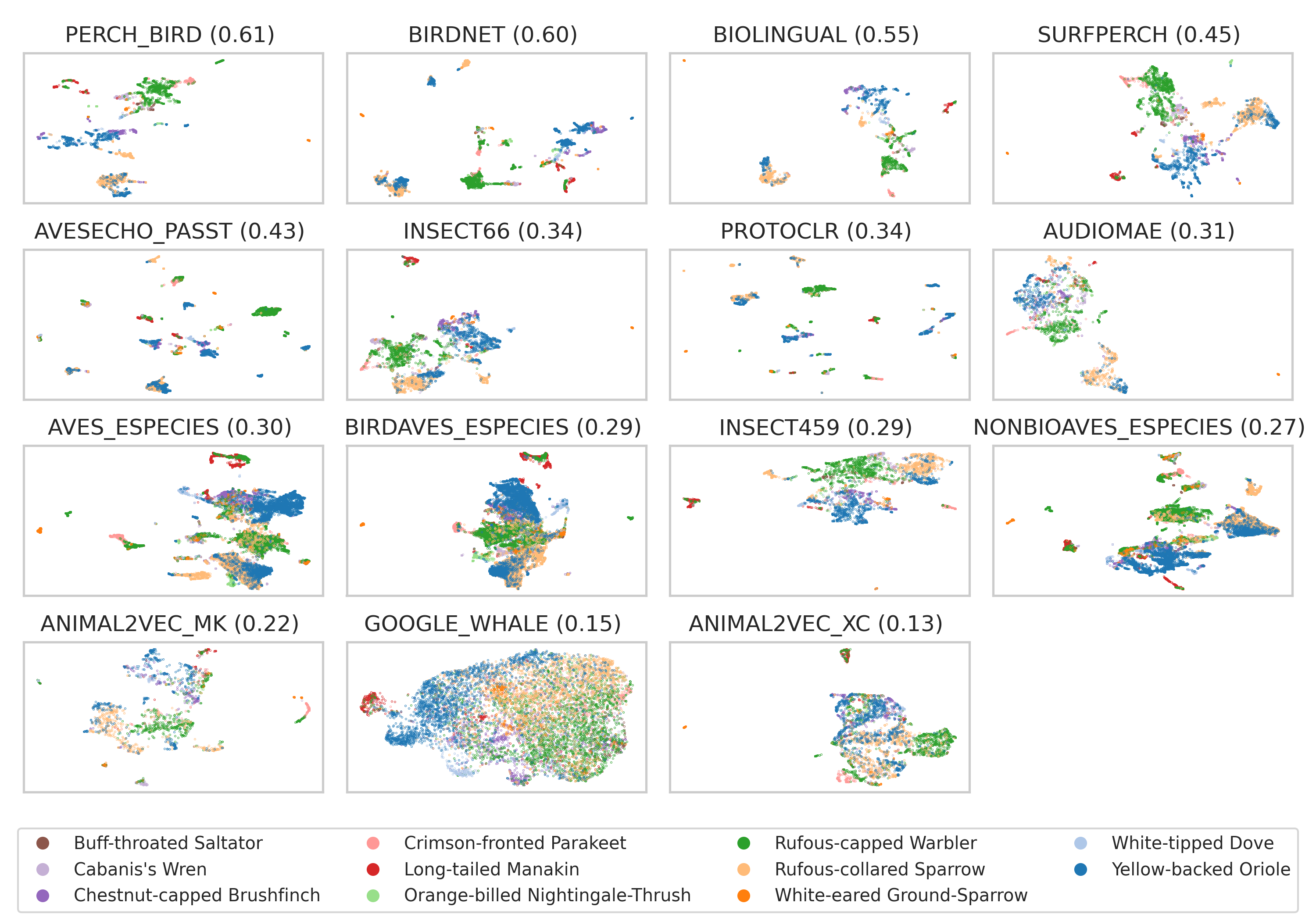}}}
    \caption{Two-dimensional embedding spaces of all feature extractors applied to bird data, sorted descending by their clustering performance of AMI values (indicated next to their name) from top left to bottom right.
    % Given the different input lengths of the feature extractors, the number of embeddings vary significantly.
    Colors correspond to the class labels, which are 11 different tropical bird species.}
    \label{fig:embeds}
\end{figure*}

For a more qualitative analysis of the different embedding spaces, two-dimensional UMAP embeddings of the feature extractors applied to the bird data are shown in Fig.~\ref{fig:embeds}.
The worst-performing feature extractors produce large unstructured clouds of mixed color, indicating that no significant clustering is achieved.
In the first and second row, feature extractors can be seen to separate the embeddings into meaningful clusters.
It is noticeable that some feature extractors such as AvesEcho\_PaSST and ProtoCLR seem to generate more subclusters than most other feature extractors.
This, however does not reflect in their performance (Fig.~\ref{fig:orig_vs_ump}).
The seven best-performing feature extractors are all trained using supervised learning and the top three are additionally trained on bird vocalizations.
All three of the AVES models (BirdAVES, AVES and NonBioAVES) reach similar performances in spite of big differences in their fine-tuning datasets \cite{hagiwara_aves_2022}.
Despite the embedding spaces of BirdAVES and Animal2vec\_XC looking similarly cluttered, their performance varies by clustering and even more so by classification (see Fig.~\ref{fig:orig_vs_ump}).

\section{Discussion}
\label{sec:discussion}

Although the self-supervised feature extractors represented in this study are trained on very large datasets, their clustering performance is inferior to most of the supervised learning feature extractors.
Nonetheless, their lack of supervision benefits generalization: the majority of self-supervised feature extractors improve performance on the frog data, both for clustering and classification.
Yet, as has been shown in previous studies \cite{ghani_global_2023}, supervised learning and training data consisting of large bird song databases yield the best results, even for challenging PAM datasets.
Perch and BirdNET \cite{kahl_birdnet_2021}, both of which are trained on thousands of classes of bird vocalizations, vastly outperform the other feature extractors by clustering and for the bird data also by classification. 
%It is worth mentioning that they do so, while being trained on standard EfficienNETs, rather than custom architectures developed for a specific task.
Despite the authors' claims \cite{robinson_transferable_2023}, the multi-modal feature extractor Biolingual, which performs very well by clustering but comparably poor by classification, is outperformed by BirdNET and Perch.
SurfPerch \cite{williams_leveraging_2024} and AvesEcho\_PaSST \cite{ghani_generalization_2024}, both of which have largely benefited from Perch or BirdNET pretraining, perform well for both clustering and classification tasks.
Most notably, AvesEcho\_PaSST is the only supervised learning feature extractor that demonstrates generalization by improving performance on the frog data.
This could be attributed to the model's innovative transformer architecture using patchout, thereby selectively dropping patches of spectrograms during training, whilst having learned from the well-trained BirdNET embeddings through knowledge distillation \cite{ghani_generalization_2024}.

% Interestingly, Insetct66 and Insect459 both of which were trained on only insect sounds and far fewer classes than the bird-trained biolingual, vastly outperform it in classification.
As stated in the introduction, self-supervised learning models lack supervision and might therefore learn classes not meaningful to differentiate between species vocalizations.
When comparing Figures \ref{fig:embeds} and \ref{fig:orig_vs_ump} we observe poor clustering both in terms of AMI performance and by qualitative visual analysis of the embedding separation, which could be resulting from non-meaningful classes.
However, for the three AVES feature extractors, the kNN classifier is nonetheless able to learn a meaningful differentiation between the classes.
This could be attributed to the fact, that while they are self-supervised, the data used to train these models consisted of curated and non-sparse sound events, thereby increasing the likelihood that meaningful classes are learned.
Furthermore, the HuBERT architecture used for the AVES models uses tokenization through acoustic unit discovery \cite{hagiwara_aves_2022}.
These acoustic units are generated during training based on 39-dimensional Mel-frequency ceptral coefficients (MFCC) features and labeled using a K-Means clustering.
The transformer encoder then predicts these cluster labels.
This differs fundamentally from the training procedure of the Animal2vec models and AudioMAE, both of which rely on masked prediction.
In terms of classification, the results suggest that the MFCC features lead to more meaningful classes than the prediction of masked emebddings.
This is supported by the drastic performance increase of all three AVES models when evaluated for the frog dataset.

For the self-supervised learning feature extractors, fine-tuning seems to only marginally improve performance.
The three feature extractors based on the AVES models all share the same general audio pretraining and architecture but differ largely in fine-tuning. 
The similarity in performance indicates that the dominant influence on the structuring of embeddings is defined by either the pretraining or the architecture.
Animal2vec\_XC and Animal2vec\_MK, the latter of which is fine-tuned, largely share the same architecture but were trained on very different datasets.
Yet, both models reach similarly poor performance, this is especially surprising for the fine-tuned Animal2vec\_MK.

%For the supervised learning feature extractor SurfPerch, which was developed for marine data in coral reefs, bird training data from Perch was mixed with coral reef sounds during pretraining.
%While the target domain is very different from the bird dataset, SurfPerch still reaches very high performance.
%This performance drops in terms of classification though, as soon as the target domain is shifted to the frog dataset.
%Again, this indicates that including data of the target domain is more effective during pretraining than fine-tuning.

While the dimensions of the feature extractors vary greatly, performance does not correlate with dimensionality.
When we used UMAP to standardize the embedding spaces, we saw performance improvements (Table~\ref{tab:results}).
The observed improvements are more pronounced in clustering, likely because UMAP is an optimization algorithm that trains a neural network to find an optimized lower dimensional graph representation,  which benefits K-Means.
%We demonstrated that using UMAP to standardize the embedding spaces generally improves performance, but it does so differently for clustering and classification (Table~\ref{tab:results}).

%Nonetheless, in Table \ref{tab:results} we demonstrated that using a standardized embedding space affects clustering and classification performance differently.
% The performance differences that can be observed are predominantly in clustering, indicating that the graph structure, which K-Means builds for the clustering, is aided by dimensionality reduction using UMAP.
% For this study dimensionality reduction using Principal Component Analysis was also performed and evaluated using linear classification, however, performance only changed marginally and was therefore omitted from this comparison.

% This analysis underlines the high quality of embeddings created by large supervised learning feature extractors like BirdNET and Perch.
Although the bird data were in-domain for BirdNET and Perch, the same applies for Biolingual, Animal2vec\_XC and BirdAVES\_ESpecies, none of which reached similar performance.
The decrease in classification performance by Perch on the frog data raises the question whether this difference can be attributed to its training dataset.
Perch's training data, which is comprised of solely xeno-canto recordings perhaps makes it less domain agnostic than BirdNET which was trained on selected curated bird datasets along with a large portion of xeno-canto.

This study presents a workflow for in-depth analysis of embedding spaces, which can be reproduced with the provided repository bacpipe.
Evaluation through clustering and classification has shown to vary significantly when applied to the different evaluation sets, undermining the use of both metrics to better understand how training setup affects performance.
While the results presented here allow for further discussion, we believe streamlining the process of evaluating and comparing such a wide variety of feature extractors can help to investigate performance differences as a function of training paradigm and training domain.
By establishing a default analysis of feature extractors alongside the common classification benchmarks, bioacoustic research can accelerate towards a better understanding of what training characteristics are beneficial in this domain.

\section{Conclusion}
\label{sec:conslusion}

In this study we have compared a variety of different state-of-the-art bioacoustic deep learning models, representing different training paradigms and training domains.
To compare the models, we have isolated their feature extractors and used them to generate embeddings of two curated evaluation datasets consisting of annotated bird and frog sounds.
The aim of this study was to firstly present a large comparison of very different bioacoustic deep learning feature extractors and to evaluate how training setup affects performance.
Performance was evaluated through clustering using an AMI score and through kNN classification using a macro accuracy score.

We have shown that bioacoustic feature extractors still struggle with polyphonic PAM datasets, especially if they are outside of the training domain.
% At this point, self-supervised learning performance is still inferior to that of supervised learning models.
However, when comparing training paradigms,  supervised learning remains a strong method to pretrain a feature extractor for general use, despite advances in self-supervised learning.
This performance difference is visible in both kNN classification and even more in clustering.
Furthermore, we have shown that alignment of training domain and target domain during pretraining impacts performance more than during fine-tuning.
This study presents a roadmap for a more in-depth performance evaluation of bioacoustic deep learning models, allowing for a better understanding of how training setup impacts downstream performance.

\section{Acknowledgments}
A lot of work went into developing each of the feature extractors presented here, and we highly appreciate that they are made publicly available.
We would also like to acknowledge Álvaro Vega-Hidalgo as well as Juan Sebastian Cañas and their coauthors, who collected, assembled and annotated the datasets used to evaluate the feature extractors. 
This project is funded by the Marie Skłodowska-Curie doctoral network BioacousticAI.
% For bibtex users:
% \bibliography{FA2025_template}
\bibliography{References}

\end{document}